\documentclass[sigconf]{acmart}

\usepackage{amsmath,amssymb,amsfonts}
\usepackage{algorithm}  
\usepackage{graphicx}
\usepackage{enumitem}
\usepackage{float}
\usepackage{algpseudocode}  
\usepackage{booktabs}
\usepackage{footnote}

\usepackage[most]{tcolorbox}
\usepackage{multirow}
\usepackage{textcomp}
\usepackage{graphicx}
\usepackage{epstopdf}
\usepackage{multirow}
\usepackage{subfigure}
\usepackage{xspace}
\usepackage{threeparttable}
\usepackage{xcolor}
\usepackage{array}
\usepackage{balance}
\usepackage{hyperref}
\usepackage{breakurl}
\usepackage{url}
\usepackage{comment}
\usepackage{alltt}
\usepackage{amsmath}
\usepackage{amssymb}
\usepackage{tabulary}
\usepackage{tikz}  
\usepackage{amsthm}
\usepackage{multirow}
\usepackage{lipsum}
\usepackage{fancyhdr}

\usepackage[most]{tcolorbox}

\usepackage[framemethod=tikz]{mdframed}

\usepackage{algorithm,algpseudocode}
\usepackage{caption}
\usepackage{setspace}

\def\BibTeX{{\rm B\kern-.05em{\sc i\kern-.025em b}\kern-.08emT\kern-.1667em\lower.7ex\hbox{E}\kern-.125emX}}
\newcommand{\nosection}[1]{\vspace{2pt}\noindent\textbf{#1.}}

\newcommand{\squishlist}{
	\begin{list}{$\bullet$}
		{ \setlength{\itemsep}{0pt}      \setlength{\parsep}{3pt}
			\setlength{\topsep}{3pt}       \setlength{\partopsep}{0pt}
			\setlength{\leftmargin}{3.5mm} \setlength{\labelwidth}{1em}
			\setlength{\labelsep}{0.5em} } }
	\newcommand{\squishend}{
\end{list}  }
\newcommand{\cut}[1]{}

\copyrightyear{2021}
\acmYear{2021}
\setcopyright{acmcopyright}\acmConference[CIKM '21]{Proceedings of the 30th ACM International Conference on Information and Knowledge Management}{November 1--5, 2021}{Virtual Event, QLD, Australia}
\acmBooktitle{Proceedings of the 30th ACM International Conference on Information and Knowledge Management (CIKM '21), November 1--5, 2021, Virtual Event, QLD, Australia}
\acmPrice{15.00}
\acmDOI{10.1145/3459637.3482361}
\acmISBN{978-1-4503-8446-9/21/11}
\begin{document}
\fancyhead{}
\title{Large-Scale Secure XGB for Vertical Federated Learning}
\thanks{$^*$Corresponding author.}

\author{Wenjing Fang, Derun Zhao, Jin Tan, Chaochao Chen$^*$, Chaofan Yu, Li Wang, Lei Wang, }
\author{Jun Zhou, Benyu Zhang}
\email{{bean.fwj, zhaoderun.zdr, tanjin.tj, chaochao.ccc, shuyan.ycf, raymond.wangl, shensi.wl, jun.zhoujun, benyu.z}@antgroup.com}
\affiliation{
  \institution{Ant Group}
}
\renewcommand{\shortauthors}{Wenjing Fang et al.}
\email{}

\begin{abstract}

Privacy-preserving machine learning has drawn increasingly attention recently, especially with kinds of privacy regulations come into force.
Under such situation, Federated Learning (FL) appears to facilitate privacy-preserving joint modeling among multiple parties. 
Although many federated algorithms have been extensively studied, there is still a lack of secure and practical gradient tree boosting models (e.g., XGB) in literature. 
In this paper, we aim to build large-scale secure XGB under vertically federated learning setting. 
We guarantee data privacy from three aspects.
Specifically, (\romannumeral1) we employ secure multi-party computation techniques to avoid leaking intermediate information during training, (\romannumeral2) we store the output model in a distributed manner in order to minimize information release, and (\romannumeral3) we provide a novel algorithm for secure XGB predict with the distributed model.
Furthermore, by proposing secure permutation protocols, we can improve the training efficiency and make the framework scale to large dataset.
We conduct extensive experiments on both public datasets and real-world datasets, and the results demonstrate that our proposed XGB models provide not only competitive accuracy but also practical performance.

\end{abstract}



\maketitle

\section{Introduction}\label{intro}
With the concern of data privacy and the requirement of better machine learning performance, 
Privacy Preserving Machine Learning (PPML) \cite{mohassel2017secureml}, which enables multiple parties jointly build machine learning models privately, has been drawing much attention recently. To date, most existing PPML models only focus on linear regression \cite{hall2011secure,gascon2017privacy}, logistic regression \cite{mohassel2017secureml,kim2018secure}, and neural network \cite{wagh2018securenn,rachuri2019trident,zheng2020industrial}. 
They are mainly divided into two types based on how data is partitioned, i.e., horizontally partitioned PPML that assumes each party has a subset of the samples with the same features, and vertically partitioned PPML that assumes each party has the same samples but different features \cite{yang2019federated}. The former usually appears when participants are individual customers (2C), while the latter is more common when participants are business organizations (2B). In this paper, we focus on the later setting. 

Gradient tree boosting is one of the widely used type of machine learning models that shine in different fields, e.g., fraud detection \cite{rushin2017horse}, recommender system \cite{zhang2018industrial}, and online advertisement \cite{ling2017model}. 
As an optimized implementation, XGB \cite{chen2016xgboost} achieves promising results in various competitions and real-world applications, since it is a gradient boosting model whose key idea is applying numerical optimization in function space and optimizing over the cost functions directly \cite{friedman2001greedy}. 
Therefore, how to build privacy preserving XGB with vertically partitioned data is an important research topic. 

So far, there are several existing researches on privacy preserving gradient tree boosting \cite{liu2019boosting,li2019practical,fengy2019securegbm,cheng2019secureboost}. Among them, \cite{liu2019boosting,li2019practical} are built for horizontally partitioned data setting. Although privacy preserving XGB are proposed for vertically partitioned data setting in \cite{fengy2019securegbm,cheng2019secureboost}, they are not provable secure. Specifically, intermediate information, e.g., the sum of the first and second order gradients for each split and the order of split gains, is revealed during model training procedure.

Building a secure XGB is challenging due to the following reasons. 
(1) \textit{Complicated computation primitives}. Unlike other machine learning models such as logistic regression which only require secure matrix \texttt{addition} and \texttt{multiplication} primitives, XGB needs additional non-linear computation primitives, e.g., \texttt{division} and \texttt{argmax}. 
(2) \textit{High memory cost}. Most models such as neural network are suitable for mini-batch training, that is, only a small batch of samples are loaded in each training iteration, and thus they do not need large memory to support large-scale datasets. In contrast, XGB uses (sampled) full-batch dataset to build trees. Therefore, how to save memory cost is the key to large-scale secure XGB.

To solve the above challenges, in this paper, we propose to build large-scale secure XGB by leveraging hybrid secure multi-party computation techniques. 
Our key idea is taking XGB as a function that two vertically partitioned parties want to compute securely, during which 
all the information is either secretly shared or encrypted, except for the trained model as the function output. 
To be specific, we first propose a secure XGB training approach, which relies on secret sharing scheme only and serves as a baseline, and we call it as SS-XGB. 
SS-XGB is provable secure but cannot scale to large datasets since it introduces redundant computations for security concern. %
We then improve the efficiency of SS-XGB by leveraging secure permutation protocol. 
We also propose two realizations of secure permutation, based on Homomorphic Encryption (HE) and Correlated Randomness (CR), and term the secure XGBs with these two secure permutation protocols as HEP-XGB and CRP-XGB, respectively. 
HEP-XGB is computational intensive which makes it a good choice under bad network condition. CRP-XGB, in contrast, needs less computation resources (e.g., CPU and memory) and enjoys an attractive performance with high bandwidth.  
%
%
After training, each party holds a partial tree structure and secretly shared leaf weights. 
Next, we present a secure XGB prediction algorithm for two parties to jointly make predictions with their local features, tree structures, and leaf weights. 
Finally, we verify our model performance by conducting experiments on public and real-world datasets and explaining our real-world applications. 
We summarize our main contributions below:
\begin{itemize} [leftmargin=*] \setlength{\itemsep}{-\itemsep}

\item We propose an end-to-end secure XGB framework by leveraging hybrid secure multi-party computation techniques, which includes model training, model storage, and model prediction. 


\item We propose two secure permutation protocols to further improve the efficiency of SS-XGB. 

\item We conduct experiments on public and real-world datasets, and the results demonstrate the effectiveness and efficiency of our proposed secure XGB algorithms. 

\end{itemize}
\section{Related Work}

\subsection{Secure Decision Tree}

\nosection{Training}
Most secure tree building (training) approaches choose the Iterative Dichotomizer 3 (ID3) \cite{quinlan1986induction} algorithm to build decision trees. 
For example, the first secure decision tree was proposed by Lindell \& Pinkas \cite{lindell2000privacy},  where they assume data are horizontally partitioned over two parties and their protocol is based on oblivious transfer \cite{naor1999oblivious} and Yao's garbled circuit  \cite{yao1986generate}. 
Later on, Hoogh et al., proposed a secret sharing based secure ID3 protocol \cite{de2014practical} which is suitable for any number of participants who horizontally partition the dataset. 
Protocols in \cite{du2002building,wang2006classification} are for secure ID3 over vertically partitioned data, assuming that all parties have the class attribute. 

\nosection{Prediction}
Another line of work is the secure prediction (scoring) of decision tree 
\cite{wu2016privately,de2017efficient,kiss2019sok}. 
They assume that there is a server who holds a decision tree and a client who provides private features. 
After the protocol, the client obtains the classification result, without learning any information about the tree beyond its size and the prediction output, and the server learns nothing about the client's input features \cite{kiss2019sok}. 
There are also several literature on securely evaluating ID3 over horizontally partitioned data for more than two parties by HE \cite{xiao2005privacy,samet2008privacy}. 


\subsection{Privacy Preserving Gradient Tree Boosting}

\nosection{Training}
Besides secure decision tree, there are also several researches on privacy preserving gradient tree boosting. 
For example, \cite{liu2019boosting} and \cite{li2019practical} proposed federated gradient boosting decision tree and federated XGB under horizontal federated learning setting, respectively. That is, they assume data samples with the same features are distributed horizontally among mobile devices or multiple parties. 
Meanwhile, the authors of \cite{fengy2019securegbm,cheng2019secureboost,tian2020federboost} proposed privacy preserving XGB for vertically partitioned data setting. 
They are the most similar work to ours in literature. 
However, the models in \cite{fengy2019securegbm,cheng2019secureboost} are not provable secure. As we described in Section \ref{intro}, intermediate information is revealed during model training procedure. 
What is more, the leaked information cannot be quantified, which may cause potential privacy leakage in practice \cite{weng2020privacy}. 
Besides, the model proposed in \cite{tian2020federboost} relies on differential privacy, which indicates a trade-off between privacy and utility. 
In contrast, in this paper, we present a provable and lossless secure XGB by leveraging secure multi-party computation techniques, including secret sharing, HE, and correlated randomness.


\nosection{Prediction}
There is also work on building secure XGB prediction models~\cite{meng2020privacy,chen2021fed}. 
For example, Meng and Feignebaum applied HE for secure XGB prediction under the same client-server mode as decision tree. 
Later on, Chen et al. proposed a HE based federated XGB inference for multiple parties \cite{chen2021fed}, which assumes that the tree structures are distributed among multiple parties.
In this paper, we propose to leverage secret sharing for secure tree prediction, which avoids time-consuming HE operations. 
\section{Preliminary}\label{secpre}



\subsection{Settings}

\nosection{Scenario setting}
Our proposal is designed for two parties (A \& B), who partition data \textit{vertically}, to jointly build secure XGB model. 
That is, A and B have aligned their common samples but they have different features. 
Formally, let $X^{M\times N}$ be the original feature matrix with $M$ and $N$ denoting the number of samples and features, respectively. 
Under vertical FL setting, $X^{M\times N}$ is distributed to A and B, A has a feature matrix $X_{A}^{M\times N_{A}}$, and B has the other feature matrix $X_{B}^{M\times N_{B}}$, where $N_{A}$ and $N_{B}$ denote the number of features at Party A and Party B, such that $N=N_{A}+N_{B}$. 
%
We also assume one of the parties (e.g., Party A) holds the label vector $Y$. 

\nosection{System setting}
Besides the above two parties (A and B), we also employ a trusted dealer (C) to generate correlated randomness, e.g., Beaver triplet \cite{beaver1991efficient}, following the \textit{MPC with pre-processing} paradigm \cite{ishai2013power,mohassel2017secureml}. 
We introduce a secure hardware (i.e., Intel SGX) to play the role of C, in order to enable more efficient and scalable MPC. 
C is only responsible for randomness generation and has no access to the inputs and outputs of A and B.
The input-independent randomness can be massively produced by C before training starts and will be consumed by A and B during the training process. 

\nosection{Model setting}
For security, we distribute the model to two parties after training. 
Both parties share the same tree structures, but none of them has the whole split information. 
Specifically, the split information on a single node will only be visible to the party who owns the corresponding feature, and will be non-visible (dummy node) to the other party. 
Moreover, both parties collaboratively store the leaf weights in secret sharing scheme. 
To this end, we can avoid potential information leakage from the trained model.

\nosection{Security setting}
Our proposal enjoys information-theoretic security and is secure in the semi-honest adversarial setting. That is, the adversary is not restricted to be a probabilistic polynomial time algorithm, and tries to infer information although it strictly obeys the execution protocol. 
This setting is adopted by most existing secure machine learning models \cite{mohassel2018aby3,mohassel2017secureml,wagh2018securenn}.

\subsection{XGB}
XGB follows the gradient boosting framework and applies numerical optimization in function space~\cite{friedman2001greedy}. 
In machine learning, the learning task is based on the definition of the loss function.
The loss function $l(y, \hat{y})$ is defined to measure the difference between the label $y$ and the estimation $\hat{y}$.

As an addictive model, XGB~\cite{chen2016xgboost} minimizes the loss sum of all the samples 
as objective iteratively.
For each tree fitting, the algorithm first calculates necessary statistics, i.e., the first-order gradient $g_i$ and second-order gradient $h_i$ of sample $i$:
\begin{equation}
g_i=\partial_{\hat{y}_i^{(t-1)}} l(y_i, \hat{y}_i^{(t-1)}), \quad h_i=\partial_{\hat{y}_i^{(t-1)}}^2 l(y_i, \hat{y}_i^{(t-1)}),
\end{equation}
where $\hat{y}_i^{(t-1)}$ is the prediction by the end of last iteration.

Then the gradient sums can be cumulated for the instance set $I_j$ on each node j:
 \begin{equation}
G_j = \sum_{i\in I_j} g_i, \quad H_j = \sum_{i\in I_j} h_i.
\end{equation}

 It takes the second-order Taylor expansion approximation of objective and adds regularization to the model.
The optimal leaf weight $w_j^\ast$ and objective $\text{obj}^\ast$ are solved with regard to the objective function:
\begin{equation}
\label{eqweight}
w_j^\ast = -\frac{G_j}{H_j+\lambda},
\end{equation}
\begin{equation}
\label{eqobj}
\text{obj}^\ast = -\frac{1}{2} \sum_{j=1}^T \frac{G_j^2}{H_j+\lambda} + \gamma T,
\end{equation}
where $\gamma$ and $\lambda$ are the regularizers for the leaf number and leaf weights respectively.
Equation (\ref{eqobj}) is applied to measure a split proposal at each node and determines the tree structure. 
With Equation (\ref{eqweight}), the leaf weights are finally acquired.

\subsection{Secret Sharing Protocols}\label{secprot}
In general, the GMW protocol~\cite{goldreich2019play,goldreich2007foundations} shares a $l$-bit value $x$ additively in the ring $\mathbb{Z}_{2^{l}}$  as the sum of two values, where $l$ is the length of the numbers and $2^l$ is the size of the computation finite field. With different $l$, GMW can work on both Boolean and Arithmetic circuits. 
%
We describe the necessary computation protocols below under the two-party setting. 
Note that these protocols can be naturally generalized to more than two parties.

\nosection{Sharing scheme} 
We denote a secretly shared value $x$ 
by angle brackets, i.e., $\langle x\rangle$. 
Specifically, it consists of two random shares, i.e., $\langle x\rangle_{0}$ and $\langle x\rangle_{1}$, which are physically distributed at parties $P_{0}$ and $P_{1}$, respectively. According to the definition of additive secret sharing,
we have $x = \langle x\rangle_{0}+\langle x\rangle_{1}~mod~2^l$. Note that we will omit $2^l$ for conciseness in the following formulas.

\nosection{Initialization} 
$Init(x)$, transforming a private input at one party
into \textit{Secret Sharing} format among two parties. 
We take a private value $x$ of $P_{0}$ for example.
Firstly, $P_{0}$ chooses a random number $r \in \mathbb{Z}_{2^{l}}$ and sends it to $P_{1}$. Then $P_{0}$ sets its local share as $\langle x\rangle_{0}=x-r$. 
Meanwhile, $P_{1}$ finishes the computation by setting $\langle x\rangle_{1}=r$. The correctness is obvious since $\langle x\rangle_{0}+\langle x\rangle_{1}=x$. The random number $r$ guarantees privacy, as a mask of $x$, and reveals nothing to $P_{1}$. 

\nosection{Reconstruction} 
$Rec(\langle x\rangle)$. After a batch of secure computations, the final result is obtained in secret sharing representation. To reconstruct it, two parties should exchange their local shares 
and sum them up.

\nosection{Addition} $\langle z\rangle = \langle x\rangle + \langle y\rangle$. $P_{i}$ locally computes $\langle z\rangle_{i} = \langle x\rangle_{i} + \langle y\rangle_{i}$ ($i\in \{0,1\}$). 
Since $\langle z\rangle_{0}+\langle z\rangle_{1}=\langle x\rangle_{0}+\langle x\rangle_{1}+\langle y\rangle_{0}+\langle y\rangle_{1}=x+y$, $(\langle z\rangle_{0}, \langle z\rangle_{1})$ is a correct secret sharing pair of the sum $z$. Similarly, $Subtraction$ protocol can be implemented with local subtraction.

\begin{figure}[t]
	\centering
	\includegraphics[width=0.4\textwidth]{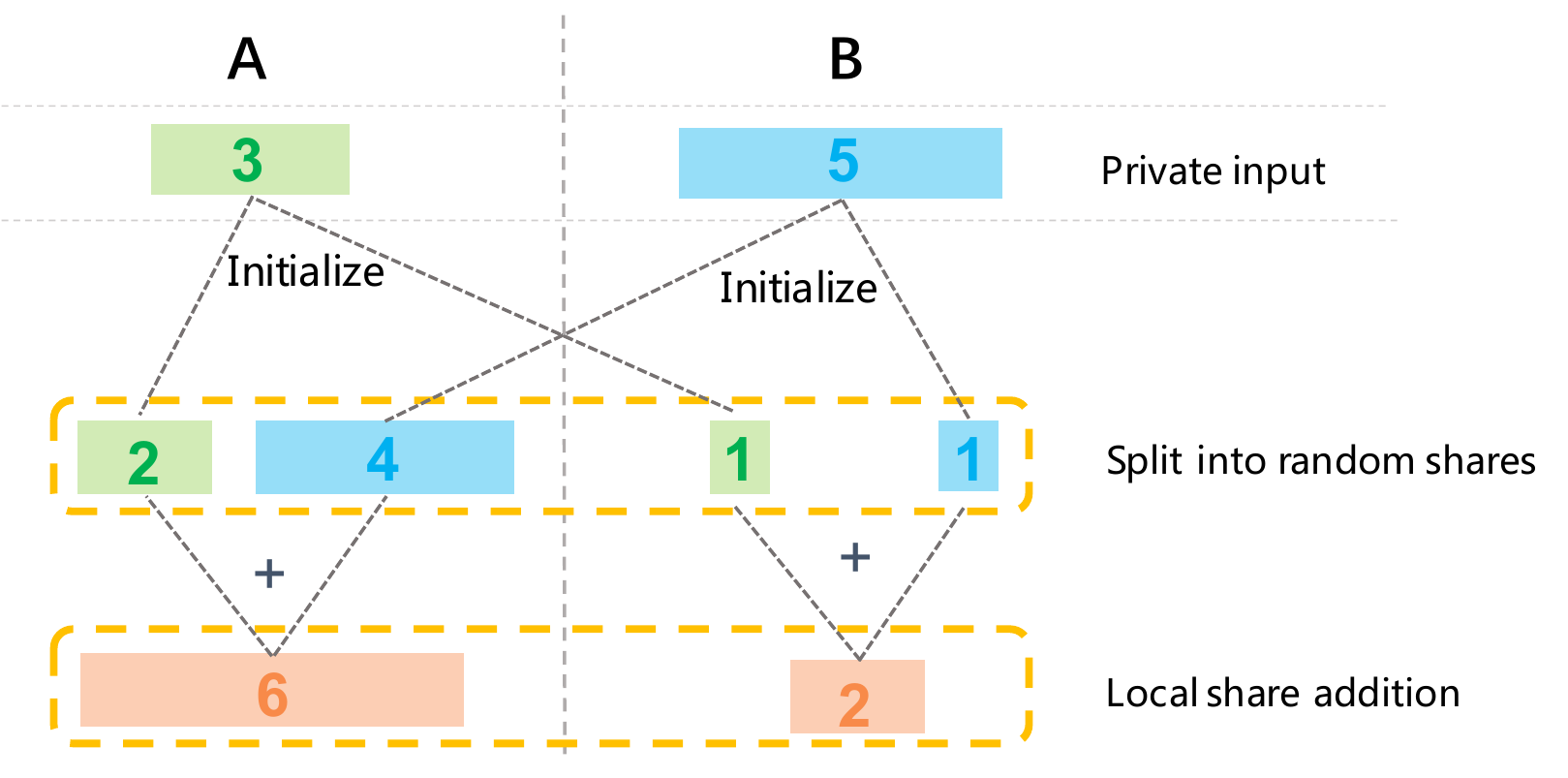}
		\vskip -0.1in
	\caption{Illustration of secret sharing addition.}\label{figsseg}
\vspace{-0.15in}
\end{figure}
	
We show an example in Figure \ref{figsseg} to better illustrate the idea of secret sharing and the above protocols. 
Two parties (A and B) hold two private values (3 and 5) separately, and want to jointly compute the sum without leaking out the original value. 
The parties first share their private inputs with the \textit{Initialization} protocol. For example, party A splits its value 3 randomly into 2 shares (2, 1) and holds one locally, i.e. number 2. 
Party B cannot infer the original value only with received share and has to guess the other share equiprobably within the finite field.
Thus the private input is protected by the share randomness.
%
Then two parties execute \textit{Addition} protocol by adding their local shares 
(e.g., $2+4=6$). 
At this point, the result shares (6, 2) make up the \textit{Secret Sharing} representation of the sum and the computation is secure. After the computation, the parties obtain the result by calling the $Reconstruction$ protocol.
%

\nosection{Scalar-multiplication} $\langle y\rangle=c \cdot \langle x\rangle$, where $c$ is the public scalar known to both parties.  $P_{i}$ locally computes $\langle y\rangle_{i}=c\cdot \langle x\rangle_{i}$ ($i\in \{0,1\}$). Then we have $\langle y\rangle_{0}+\langle y\rangle_{1}=c \cdot (\langle x\rangle_{0}+\langle x\rangle_{1})=c\cdot x$, and $(\langle y\rangle_{0}, \langle y\rangle_{1})$ is a correct secret sharing pair of the product $y$. 

\nosection{Multiplication} 
$\langle z\rangle=\langle x\rangle \cdot \langle y\rangle$. Because GMW scheme holds the additive homomorphism~\cite{evans2017pragmatic}, it can integrate with the Beaver-triplet approach to achieve an efficient online computation~\cite{beaver1991efficient}. 
A Beaver-triplet contains three secret shared values $( \langle a\rangle, \langle b\rangle, \langle c\rangle)$, where a and b are uniformly random values and $c = a\cdot b$. Then the parties compute $\langle e\rangle=\langle x\rangle-\langle a\rangle$ and $\langle f\rangle=\langle y\rangle-\langle b\rangle$ with $Subtraction$ protocol. In this way, $x$ and $y$ are securely masked by random values. $e$ and $f$ can be reconstructed without leakage. $P_{i}$ finishes the computation by computing locally $\langle z\rangle_{i}=-i\cdot e\cdot f+f\cdot \langle x\rangle_{i}+e\cdot \langle y\rangle_{i}+\langle c\rangle_{i}$.

Due to the fixed-point representation, a truncation is required as the postprocessing after each \textit{Multiplication}.
We adopt the efficient proposal in \cite{mohassel2017secureml}, which truncates the product shares independently. It has been proven that, with high probability, the reconstructed product is at most 1 bit off in the least significant position of the fractional part. This small truncation error is generally tolerable for machine learning applications with sufficient fractional bits. 

\nosection{Division} 
$\langle z\rangle=\langle x\rangle / \langle y\rangle$. We apply the Goldschmidt’s series expansion \cite{goldschmidt1964applications} and execute this numerical optimization algorithm with \textit{Addition} and \textit{Multiplication}. Following the work in \cite{catrina2010secure}, secure division offers sufficient accuracy with linear-approximated initialization and two computation iterations.

\nosection{Sigmoid} 
Classical sigmoid approximations includes Taylor expansion \cite{kim2018secure} and piece-wise approximation~\cite{mohassel2017secureml}. Taylor expansion explodes with large argument number which is a common case in large-scale training tasks. Piece-wise approximation faces the problem of precision loss, which is critical for the model accuracy. As an alternative, we apply $f(x)=\frac{0.5x}{1+|x|}+0.5$ to obtain a better approximation.

\nosection{Argmax} 
We first introduce the Boolean sharing \cite{demmler2015aby} to implement the \textit{Comparison} protocol, which enables bit decomposition of the subtraction of two numbers. In this way, we can get the sign of the subtraction which indicates the comparison result. Then, tree-reduction algorithm \cite{mohassel2020practical} computes \textit{Argmax} by partitioning the input into two halves and reduces the number of comparisons by half in each round.

\begin{algorithm}[t] 
	\begin{algorithmic}[1]  
		\caption{Training Algorithm of XGB}
		\label{alg:trainxgb}  
		\Function {TrainXGB}{$ X, Y, T, N$}
		\State $f_{0}=ComputeBaseScore(Y)$
		\State $model=[f_{0}]$  
		\For{$t=1,2,...,T$}
		\State $f_{t}=InitTree()$
		\State $P=PredictScore(X, model)$
		\State $\hat{Y}=Sigmoid(P)$
		\State $G_{t-1}=ComputeGradient(Y, \hat{Y})$
		\State $H_{t-1}=ComputeHessian(Y, \hat{Y})$
		\For{$n=1,2,...,N$}
		\If{n is a leaf}
		\State $w=ComputeWeight(G_{t-1}, H_{t-1}, f_{t}, n)$
		\State $f_{t}.SetWeight(w, n)$
		\Else
		\State $SG, SH=SumGradients(X, G_{t-1}, H_{t-1}, f_{t}, n)$
		\State $GA=ComputeGain(SG, SH)$
		\State $index=Argmax(GA)$
		\State $feat, val=RecSplitInfo(index)$
		\State $f_{t}.SetSplit(feat, val, n)$
		\State $f_{t}.SplitNode(feat, val)$
		\EndIf
		\EndFor
		\State $model.append(f_{t})$
		\EndFor
		\State \Return $model$
		\EndFunction
	\end{algorithmic}  
\end{algorithm}

\section{The proposed model}\label{secxgb}

In this section, we first present SS-XGB by leveraging secret sharing. We then propose HEP-XGB and CRP-XGB which improve the efficiency of SS-XGB by secure permutation. We finally propose a secure XGB prediction algorithm.  

%

\subsection{SS-XGB}
%

We start from the general XGB training in Algorithm \ref{alg:trainxgb} and later will turn it into a secure version with proper modifications. The inputs are a feature matrix $X$ and a label vector $Y$. As an additive model, XGB first makes an initial guess (line 2), followed by the predictions from the $T$ tree models as revisions (line 4-24), where $T$ is the number of trees. XGB simplifies the objective function with second-order Taylor approximation. Before each tree fitting process, the gradients are computed (line 8-9) with the up-to-date predictions (line 6-7). With the help of the gradients, the algorithm proceeds to split each node (line 10-22), where $N$ is the number of nodes. If we reach a leaf node, the associated weight is computed and stored in the model (line 11-13). Otherwise, we should enumerate all possible split candidates and 
compute the gradient sums of the containing instances in each split bucket (line 15). With the accumulated gradients and Equation (\ref{eqobj}), the feature and value with the max split gain are obtained (line 16-18). Then we update the split information in tree and split the instances into two child nodes (line 19-20). After traversing all the nodes, a new tree is built and appended to the model list (line 23).
In this way, the model is finally returned with a base score and $T$ trees.

Our proposed SS-XGB model remains unchanged as XGB except for the following adaptations:

\nosection{Hide source data} 
In order to protect the private inputs, the label $Y$ and the intermediate calculations, e.g., $G$, $H$, and $GA$ (line 16), are represented in secret sharing scheme. 
The computations are all replaced with the secure protocols described in Section \ref{secprot}. Following the paradigm of secret sharing, nothing but the output model will be revealed.

\begin{figure}[t]
	\centering
	\includegraphics[width=0.32\textwidth]{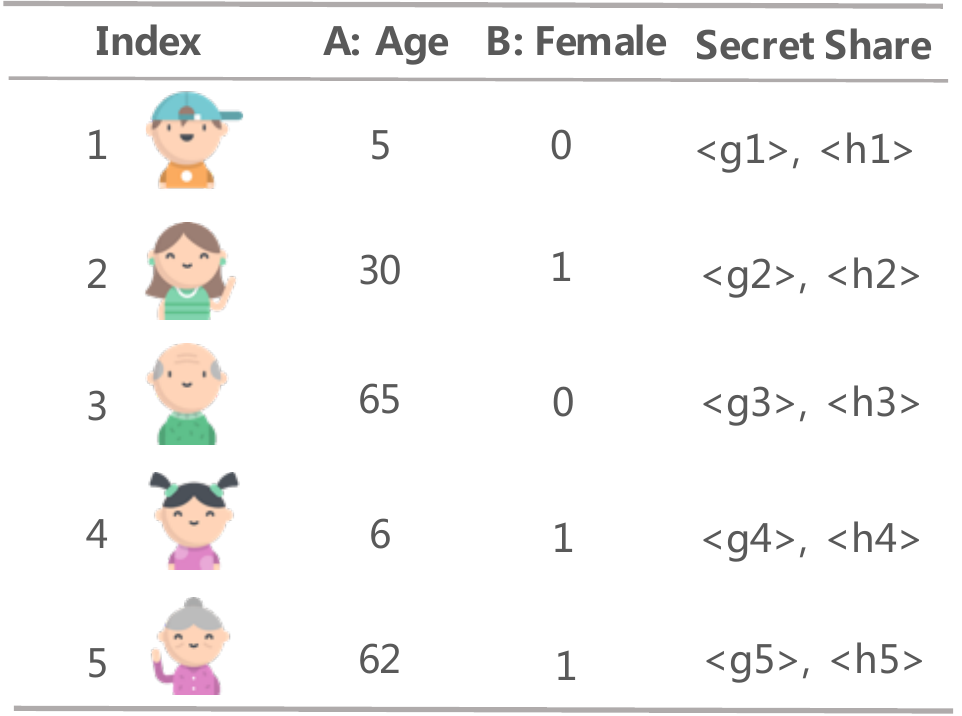}
		\vskip -0.1in
	\caption{A toy data example.}\label{figinst}
\end{figure}

\begin{figure}[t]
	\centering
	\includegraphics[width=0.4\textwidth]{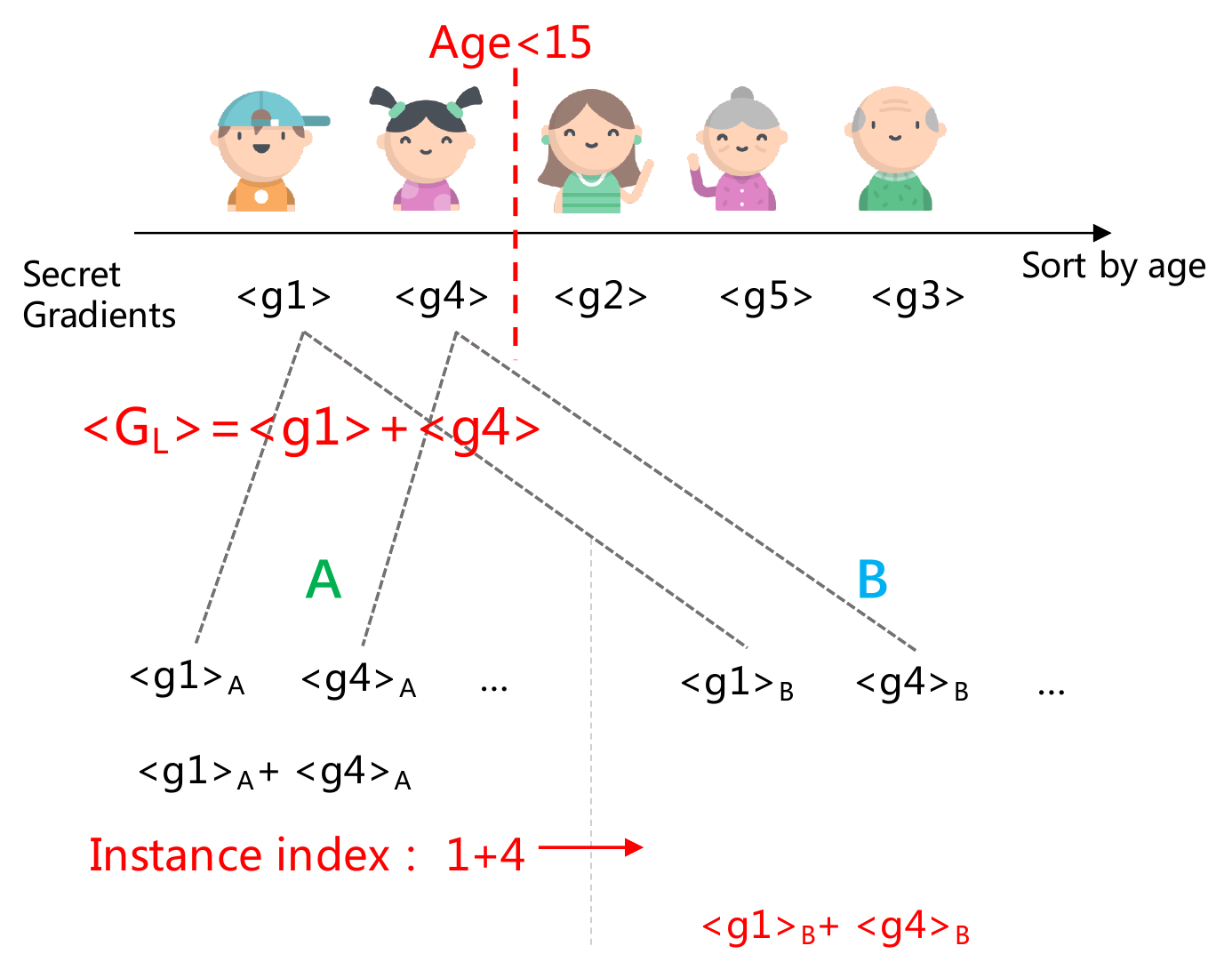}
		\vskip -0.15in
	\caption{Unsecure \textit{SumGradients}.}\label{figgsum}
\end{figure}

\nosection{Hide partial order} 
\textit{SumGradients} in Algorithm \ref{alg:trainxgb} turns out to be a major obstacle. We illustrate the problem with an example in Figure \ref{figinst}, where there are five instances in the data and the gradients are secret share values. Supposing that Party A holds a feature `Age', Figure \ref{figgsum} displays how to implement the \textit{SumGradients} directly, where we only consider the first-order gradient for convenience. 

Party A firstly sort the instances by feature `Age' and sums the gradients for all split candidates. For example, when the split condition is `Age<15', we should sum the gradients on the left side of the split point, i.e., $G_L=\langle g_1\rangle+\langle g_4\rangle$. According to the \textit{Addition} protocol in Section \ref{secprot}, Party A has to send the instance indexes to Party B, so that Party B is able to execute the local addition for correct instances.

\begin{figure}[b]
	\centering
	\includegraphics[width=0.35\textwidth]{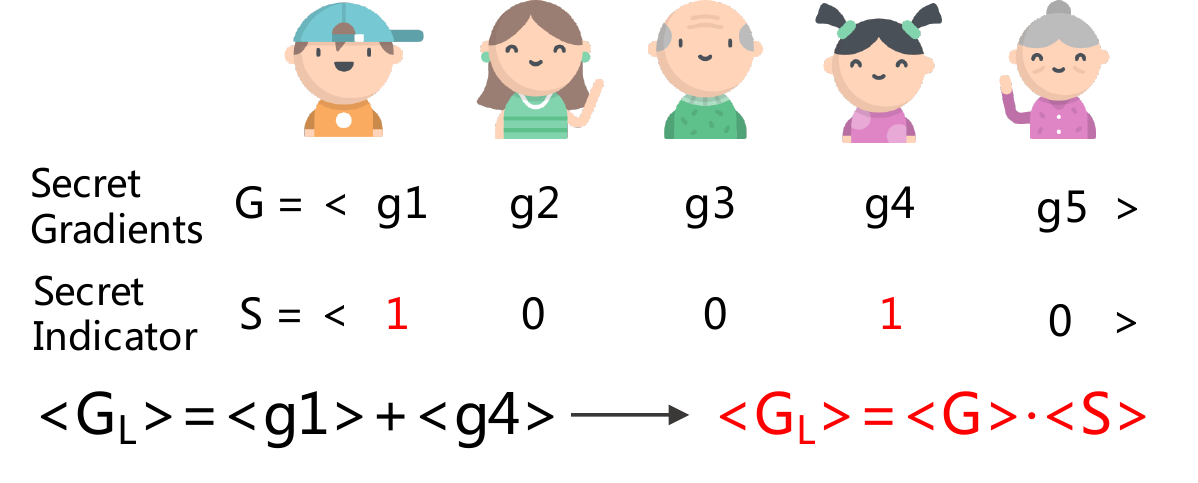}
		\vskip -0.15in
	\caption{Indicator-based secure \textit{SumGradients}.}\label{figindi}
\end{figure}

However, in the process, Party A has to leak the partial order of feature `Age' to Party B after enumerating all the split points, i.e., 1 $\rightarrow$ 4 $\rightarrow$ 2 $\rightarrow$ 5 $\rightarrow$ 3. In order to hide the order, we introduce an indicator vector (vector $S$ in Figure \ref{figindi}). The instances to be accumulated is marked as 1 in $S$ by Party A and 0 otherwise. To preserve privacy, we also transform the indicator vector to secret shares. In this way, the sum of the gradients of the instances can be computed as the inner product of the indicator vector and the gradient vector, which can be securely computed by \textit{Addition} and \textit{Multiplication} protocols.

\nosection{Hide instance distribution} 
In XGB, the instances are partitioned into two child nodes after the best split point is determined. 
The instance distribution on nodes is also intermediate information, which should not be revealed. Again, we solve this with the indicator trick. We partition the instances by updating the gradients as secretly shared zeros for the unrelated instances. 

\begin{figure}[t]
	\centering
	\includegraphics[width=\columnwidth]{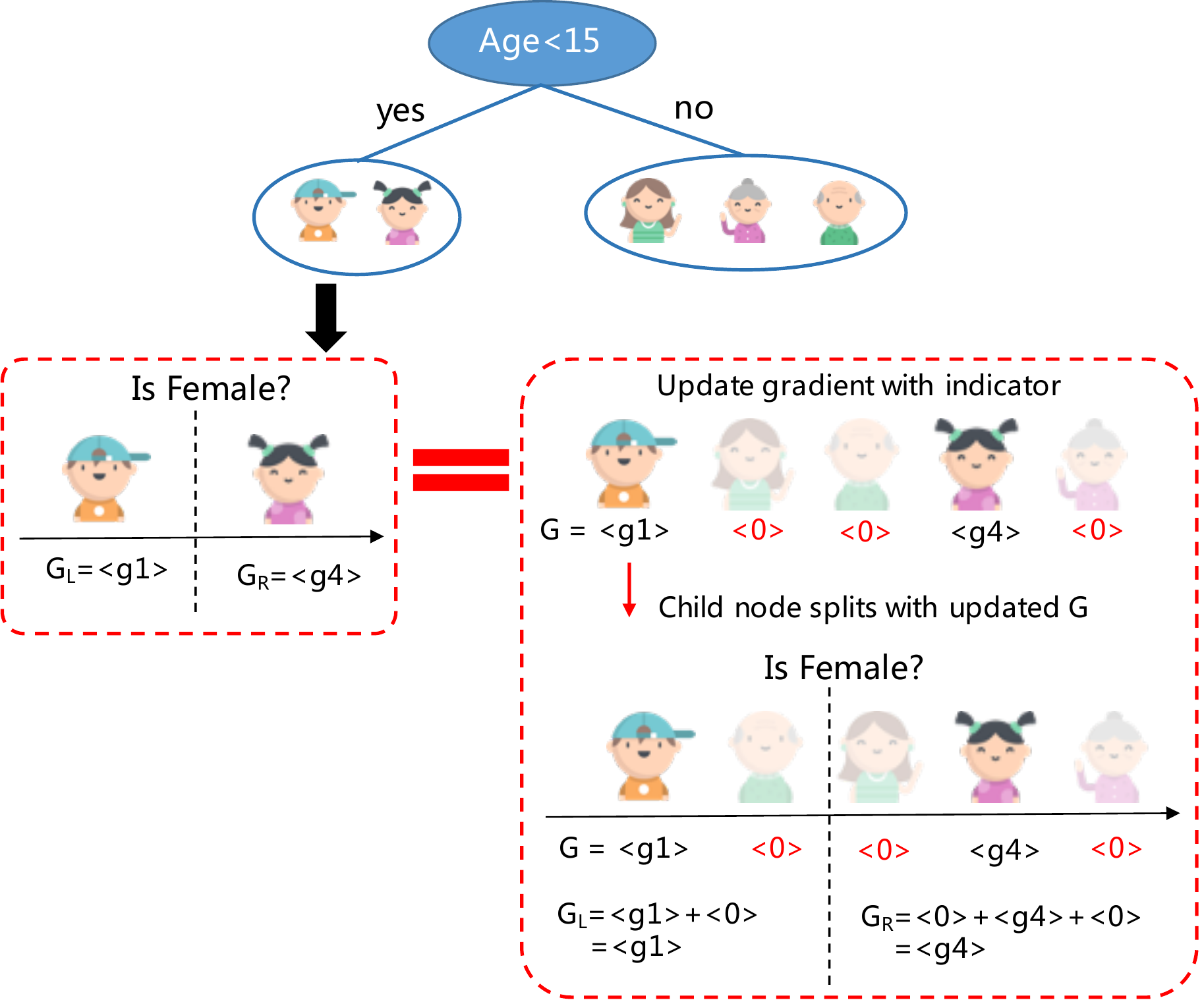}
		\vskip -0.1in
	\caption{Hide instance distribution.}\label{figdist}
\end{figure}

We illustrate why the secret partition works in Figure \ref{figdist}. 
Assuming `Age<15' is the best split point, we present how to update the gradient vector for left branch with $G_L$ in Figure \ref{figindi}. 
As we can see, the gradients for the 3 instances outside the left branch, i.e., in the right branch, are set as secret shared zeros. 
After it, we continue split the left branch with feature `Female' which is owned by Party B. 
After sorting the entire instances, the gradients on the left side of the split point can be computed. Because the invalid gradients are zeros, the accumulated gradients are equivalent to \textit{SumGradients} in the plaintext. 

With the above adaptations, two parties are able to jointly train secure XGB without any information leakage.

\subsection{HEP-XGB and CRP-XGB: Improving SS-XGB by Secure Permutation}\label{secsp}
Although the above SS-XGB is secure, it could hardly handle large-scale datasets. 
Because we introduce redundant zeros and related computations, in the indicator trick, to confuse the adversary. 
Thus, the security is achieved at the cost of the heavy communication. Supposing that there are $M$ samples and $N$ features in the dataset, if we propose $K$ split values for each feature, the indicator matrix has $M$ rows and $(N\times K)$ columns. The gradient matrix has 2 rows (for first-order and second-order gradient respectively) and $M$ columns. Then the \textit{SumGradients} turns out to be a matrix multiplication of the gradient matrix and indicator matrix. The communication complexity is the sum of two matrices, i.e., $(MNK+2M)$, which is intolerable under real-world scenarios.

\begin{figure}[b]
	\centering
	\includegraphics[width=0.25\textwidth]{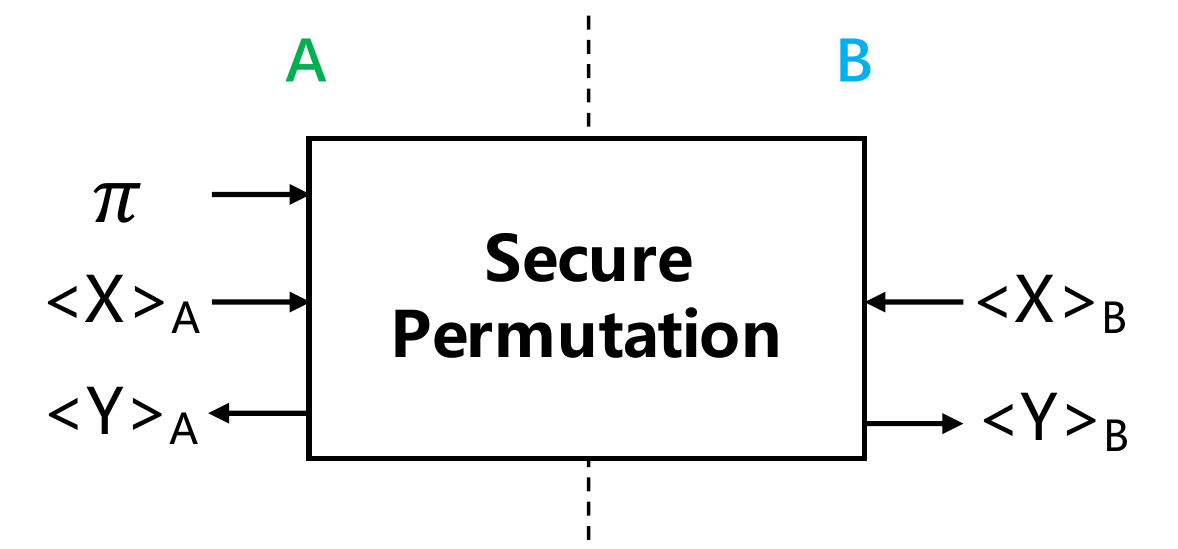}
		\vskip -0.15in
	\caption{Secure permutation. }\label{figsp}
\end{figure}

The above bottleneck of SS-XGB motivates us to build a building block which could \textit{permute a secret shared vector with a given rank}. Taking the case in Figure \ref{figgsum} for instance, we have to sort the secret gradients with respect to feature `Age'. Party A generates the mapping permutation (1, 3, 5, 2, 4), which indicates the new indexes of the original elements. 

We display the functionality of the \textit{Secure Permutation} in Figure \ref{figsp}. This block permutes a secret shared vector $\langle X\rangle$ which has $n$ elements. 
The permutation is defined as a function $\pi$, which is a private input of Party A. The output $\langle Y\rangle$ is the secret shared vector that applies $\pi$ on vector $X$, i.e., $\langle Y\rangle=\langle\pi(X)\rangle$.
Next, we will propose two implementations for this building block and compare their performance theoretically.

\nosection{HE based permutation} 
The first method is based on Homomorphic Encryption (HE). Additive HE is a form of encryption with an additional evaluation capability for computing over encrypted data without access to the secret key. It is widely used for privacy-preserving outsourced computation \cite{fun2016survey}. The typical computation scheme is regarding one party as an encrypter and the other as a calculator. All computations are conducted by the calculator with ciphertexts. Because the calculator has no access to the secret key of encrypter, the original data and computations are secure. 

Intuitively, we can take advantage of HE scheme and take the permutation provider as the calculator. 
This can be done by mainly three steps. That is, (1) A and B transform the gradient vector from secret sharing scheme to HE scheme (\textit{S2H}), after which A holds a ciphertext vector, (2) A makes permutation on the ciphertext vector, and (3) A and B transform the permuted ciphertext vector from HE scheme to secret sharing scheme (\textit{H2S}). 
In the following, we will elaborate how to securely transform between two schemes, i.e., \textit{S2H} and \textit{H2S}.

\begin{figure}[t]
	\centering
	\includegraphics[width=0.5\textwidth]{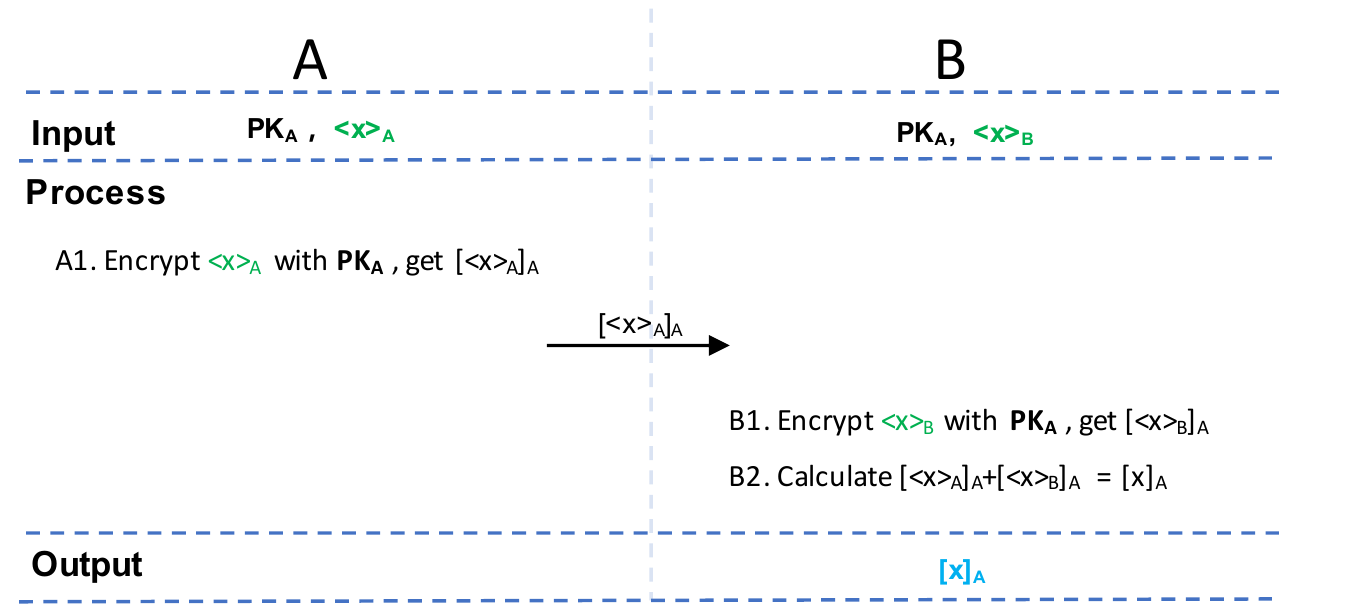}
	\caption{S2H protocol.}\label{figs2h}
\end{figure}

\textit{S2H}: 
we first represent the encrypted variable under HE scheme as $[x]_A$, where $x$ is the plaintext and $A$ denotes the encrypter. The \textit{S2H} protocol is described in Figure \ref{figs2h}, where Party A acts as the encrypter and provides the public key and Party B receives the encrypted share and outputs the sum in cipertext. In this process, Party B has no access to $SK_{A}$, and thus the security of the original value $x$ is guarantee by HE. 

\begin{figure}[t]
	\centering
	\includegraphics[width=0.5\textwidth]{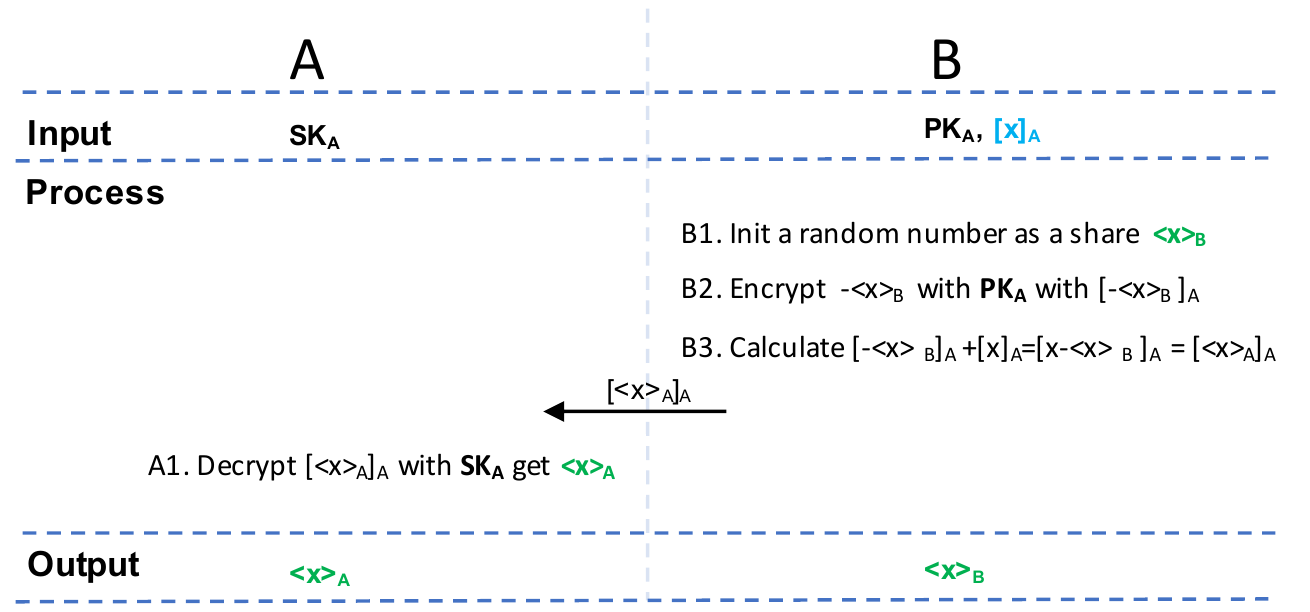}
	\caption{H2S protocol.}\label{figh2s}
		\vskip -0.1in
\end{figure}

\textit{H2S}: 
We then explain the \textit{H2S} protocol step by step in Figure \ref{figh2s} whose correctness is obvious. As for the security, we look into it from two aspects.
On the one hand, $x$ is encrypted by $PK_{A}$ and party B cannot decrypt the ciphertext. On the other hand, Party A only gets ciphertext of the random share. 
Both parties have no access to the original value, thus the transformation is secure. 

In order to reduce the communication, we accumulate the gradients under HE scheme and postpone the execution of \textit{H2S} protocol to the end of the \textit{SumGradients}. The communication is reduced to $2NK$, which is the size of the result matrix of \textit{SumGradients}.

\begin{figure}[t]
		\centering
		\includegraphics[width=0.5\textwidth]{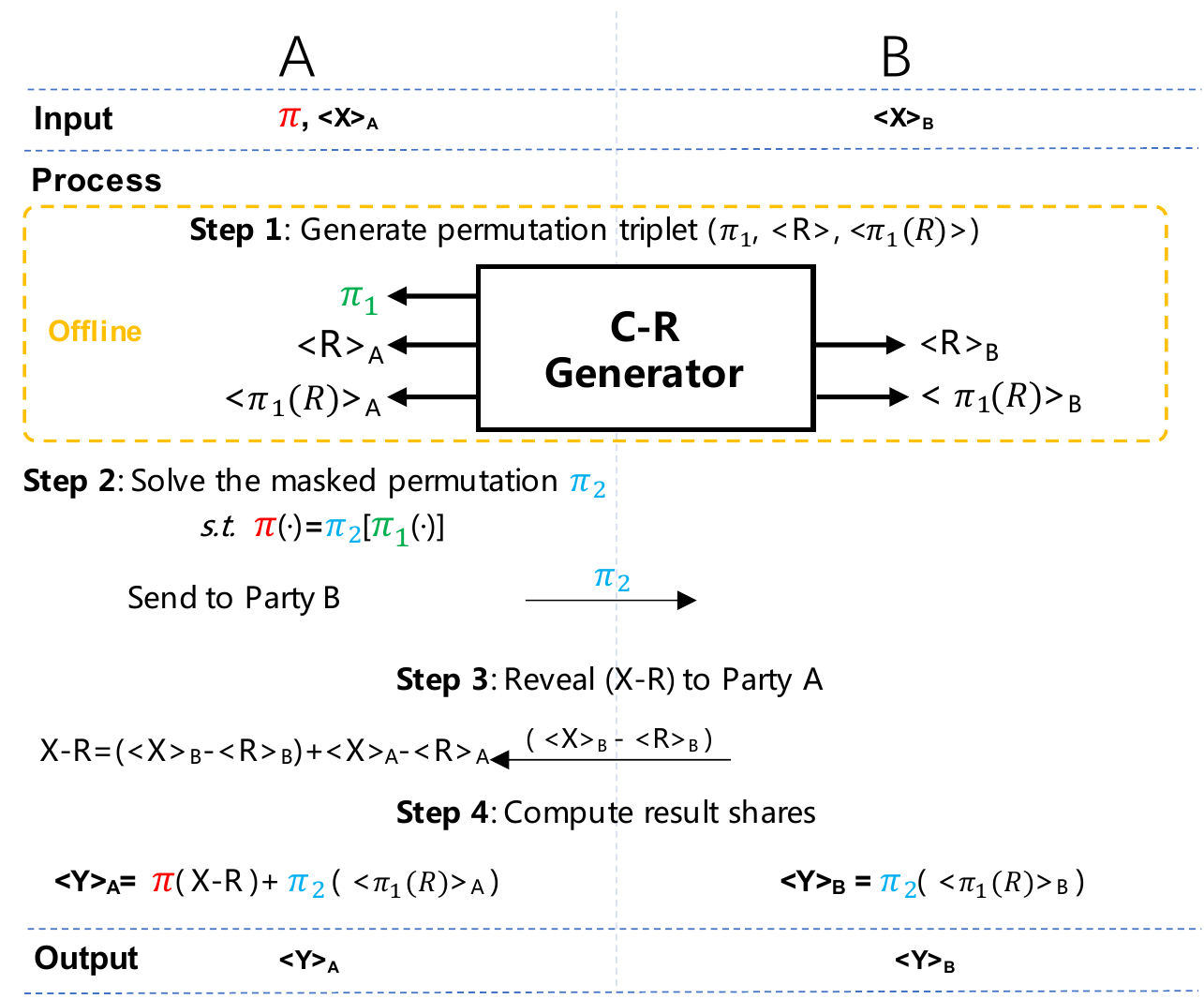}
		\caption{CR based secure permutation.}\label{figsp_prot}
\end{figure}

\nosection{CR based permutation} 
The second method relies on Correlated Randomness (CR) under the secret share scheme. 
%
%
The very first work on CR is the Beaver's multiplication triplet described in Section \ref{secprot}. The Beaver triples are input-independent random numbers and generated in the offline process. The three numbers in a triplet are correlated with each other. By utilizing this correlated randomness, the online process obtains significant speedup. 
There are also other types of CR, including garbled circuit correlation, OT correlation, OLE correlation, and one-time truth-tables~\cite{ishai2013power,damgaard2017tinytable,couteau2019note}.
Similarly, we can design specific CR for permutation function to achieve better performance.
We elaborate the CR based permutation protocol in Figure \ref{figsp_prot}, which mainly has four steps. 

In step 1, we employ the trust dealer to generate the triplet and dispatch the secret shares to both parties. We define the permutation triplet as ($\pi_{1}, \langle R\rangle, \langle  \pi_{1}(R)\rangle$), where $\pi_{1}$ is a random permutation order, $R$ is a vector with $n$ random numbers, and $\pi_{1}(R)$ denotes the result vector after permuting $R$ with $\pi_{1}$.  This step is independent of the XGB training process, and thus can be executed in the offline phase.

In step 2, we solve the transformation permutation function $\pi_{2}$, using the source permutation $\pi_{1}$ and the target permutation $\pi$. To do this, Party A performs this computation and sends $\pi_{2}$ to Party B. Note that, $\pi_{1}$ works as a random mask for $\pi$. Party B only receives the masked permutation $\pi_{2}$ and thus $\pi$ is securely hided.

In step 3, a $Reconstruction$ protocol is executed and reveals $(X-R)$ to Party A.
Similarly, the private vector $X$ is masked by the random vector $R$ in an element-wise manner. Thus, the $Reconstruction$ leaks out nothing about $X$.

In step 4, the parties locally compute the output shares. One can check the correctness as follows:
\begin{equation}
\begin{split}
Y &= \langle Y\rangle_{A}+\langle Y\rangle_{B} \\
&=\pi(X-R)+\pi_{2}[ \langle  \pi_{1}(R)\rangle_{A}+ \langle  \pi_{1}(R)\rangle_{B}] \\
&=\pi(X-R)+\pi_{2}[\pi_{1}(R)] \\
&=\pi(X-R)+\pi(R) =\pi(X).
\end{split}
\end{equation}

The secure permutation protocol is employed to sort the first-order and second-order gradients for XGB. The \textit{SumGradients} then accumulates the gradients with \textit{Addition} protocol, which contains only cheap local computations.
%
We term the two optimized approaches, based on HE and CR secure permutation implementations, as \textbf{HEP-XGB} and \textbf{CRP-XGB} respectively.

\nosection{Complexity analysis} 
In general, SS-XGB and CRP-XGB only rely on secret sharing scheme, and thus they are computational efficient but have high communication cost. 
In contrast, since it is time-consuming for HE computations,  HEP-XGB is computationally intensive but with less communication overhead. 
All these approaches share the same algorithm flow except for \textit{SumGradients}, we summarize their time costs in this process below:
\begin{itemize}[leftmargin=*]
    \item SS-XGB: the communication is as analyzed in section \ref{secsp}, i.e., $MNK+2M\approx MNK$. 
    \item CRP-XGB: to accumulate first and second order gradients, the online phase transmits a masked permutation and a revealing share, thus the total communication is $2MN$ plaintexts and $2MN$ shares. Compared with SS-XGB, CRP-XGB removes the bucket number ($K$) which is set to 33 by default. Therefore, not only the communication but also the memory are significantly reduced.
    \item HEP-XGB: the \textit{S2H} protocol is invoked for $2M$ times to transform gradients. And we need $2MN$ homomorphic additions to accumulate the encrypt gradients to belonging bucket, and $2KN$ invocations for \textit{H2S} protocol to transform the accumulated result back to secret sharing scheme. In total, it takes $2M$ \textit{Encryption}s, $2MN$ \textit{Addition}s, and $2KN$ \textit{Decryption}s under homomorphic encryption scheme. And the two transformation protocols need to communicate $(2M+2NK)$ ciphertexts, which is much longer than secret shares.
\end{itemize}

We will experimentally compare CRP-XGB and HEP-XGB and describe how to choose a proper one in Section \ref{secexp}.

\subsection{Secure XGB Prediction Algorithm}
\begin{figure}[htbp]
		\centering
		\includegraphics[width=0.43\textwidth]{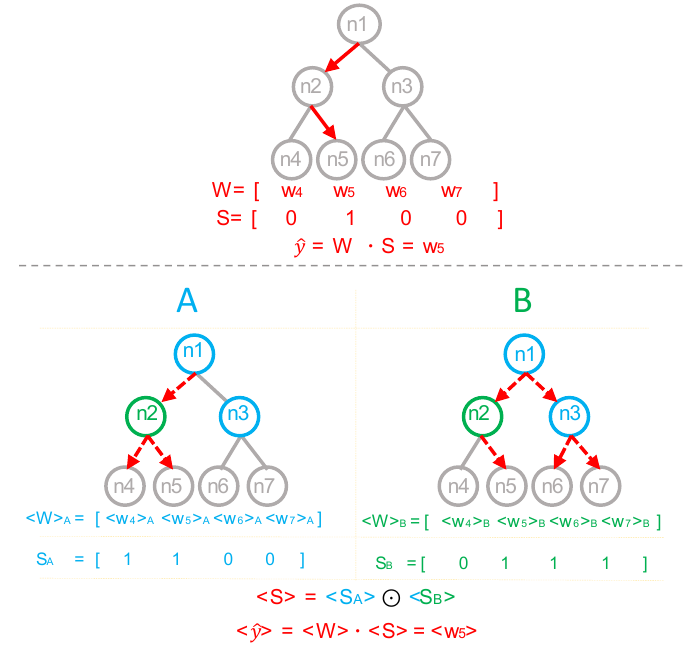}
		\vskip -0.05in
		\caption{Illustration of secure tree prediction}\label{figpred}
\end{figure}

We present a novel secure XGB prediction algorithm with the example in Figure \ref{figpred}. The upper part
displays the normal prediction process over a single tree. Prediction starts from
the root node and recursively chooses a child node determined by the split information.
Moving along the prediction path (shown in red arrows) until the leaf ($n_{1}\rightarrow n_{2}\rightarrow n_{5}$),
the corresponding weight ($w_{5}$) is finally returned
as the prediction. To facilitate the secure version, we introduce an one-hot indicator vector $S$ to mark the selected leaf node. The prediction could be regarded as
the inner product of S and W (a vector consists of all leaf weights).

Supposing that the intermediate nodes $n_{1}$ \& $n_{3}$ are split by
party A and $n_{2}$ is a dummy node for party A. On the contrary, party B is
only aware of the split information of $n_{2}$. Leaf weights are variables under
SS domain, each party holds one of the shares. The output model of our secure
training algorithm is distributed just as we marked in the bottom part of
Figure \ref{figpred}. 
It explains an innovative way to achieve
similar prediction process under the SS domain. Both parties first
generate leaf indicator vector locally. 
For each party (e.g., A), if its intermediate node has split
information (e.g. $n_{1}$ of party A), the unselected branch is abandoned (e.g. $n_{3}$ of party A). 
Otherwise, prediction continues on both branches (e.g. $n_{2}$ of party A).
The local indicator vector marks all the candidate leaves with the partial model.
Obviously, the intersection of local indicator vectors is equivalent to the
one-hot indicator vector $S$. So we transform the local indicator vector to
SS domain and get $S$ with an element-wise multiplication. The prediction
now can be computed efficiently with the weight vector and the indicator vector under SS domain.
By cumulating predictions of all the trees, we obtain the final prediction of XGB. 

\section{Experiments}\label{secexp}

In this section, we present the comprehensive experiments we conducted to study the effectiveness and efficiency of our proposed secure XGB algorithms.

\subsection{Experimental Setup}\label{secsetup}
\nosection{Implementation}
We use $k=128$ bits to represent the fixed-point secret sharing scheme, and 20 out of which are used for the fractional part. We use 64 bits to represent the plaintext data. Zero sharing technique is applied to reduce the interaction cost with the help of pseudo-random functions (PRF) \cite{mohassel2018aby3}. 
As for homomorphic encryption, we implement the Okamoto-Uchiyama scheme~\cite{okamoto1998new} based on \textit{libtomath}\footnote{https://github.com/libtom/libtommath} and the key size is set to 2,048 bit. 
HEP-XGB needs the transformation protocols and we have to deal with the security concern of random number generation. Following the analysis in \cite{pullonen2012design}, we pick the random number $r \in \mathbb{Z}_{2^{k+\sigma}}$. The probability of information leaking is then reduced to $2^{-\sigma}$, where we set $\sigma =40$.

\nosection{Environment}
We conduct experiments on the machines equipped with Intel(R) Xeon(R) Platinum 8,269CY CPU @ 2.50GHz$\times$32 and 128G of RAM. We simulate multiple parties by initializing multiple dockers with the given core limit. The network condition is manipulated with the traffic control command of Linux. To be specific, we consider the transmission $bandwidth$ and $latency$.

\nosection{Datasets}
We choose two types of datasets, which belong to different scales. 
First, we use four \textbf{public} benchmark datasets, including  Energy\footnote{https://www.kaggle.com/loveall/appliances-energy-prediction}  (regression task with 19,735 samples and 27 features), Blog\footnote{http://archive.ics.uci.edu/ml/datasets/BlogFeedback} (regression task with 60,021 samples and 280 features), Bank\footnote{http://archive.ics.uci.edu/ml/datasets/Bank+Marketing\#} (classification task with 40,787 samples and 48 features after preprocessing), and Credit\footnote{https://www.kaggle.com/uciml/default-of-credit-card-clients-dataset} (classification task with 30,000 samples and 23 features). The features are evenly divided into two parties. 
We split 80\% of the samples as training dataset, unless the division has been given. In terms of efficiency test, we generate synthetic data with \textit{sklearn}\footnote{https://scikit-learn.org/stable/} toolkit. 
Second, we use two \textbf{real-world} datasets, including a dataset for regression (617,248 samples for training and 264,536 samples for testing) and another dataset for classification (140,410 samples for training and 111,618 samples for testing). 
For the regression dataset, Party A holds 25 features and label, and Party B holds 29 features. For the classification dataset Party A holds 16 features and label, and Party B holds 7 features.

\nosection{Evaluation metrics}
For the four public datasets, we use Area Under the ROC Curve (AUC) as the evaluation metric for classification tasks, and use Root Mean Square Error (RMSE) as the metric for regression tasks.
For the two real-world datasets, besides AUC and RMSE, we also use four commonly used metrics in industry, including F1 score\footnote{https://en.wikipedia.org/wiki/F-score}, KS value\footnote{https://en.wikipedia.org/wiki/Kolmogorov\%E2\%80\%93Smirnov\_test}, coefficient of determination\footnote{https://en.wikipedia.org/wiki/Coefficient\_of\_determination} (R$^2$), and Mean Absolute Error (MAE). 
F1 and KS are used for classification task, while R$^2$ and MAE are used for regression task.

\subsection{Accuracy Comparison on Public Datasets}\label{secacc}

We compare accuracy with the plaintext XGB on the above four public datasets. During experiment, all the models share the same parameters. We set $max\_depth=5$ and the \textit{tree\_method} as `approx' in order to enable the quantile sketch splitting. We set the tree number as 30 and 20 for regression tasks and classification tasks, respectively. We repeat each experiment 6 times and report the average results in Table \ref{tab:acc}.

\begin{table}[t]
  \centering
  \caption{Accuracy comparison on public datasets.}
  \label{tab:acc}
    \begin{tabular}{lllll}
    \toprule
    \textbf{Dataset} & \textbf{XGB} & \textbf{SS-XGB} & \textbf{HEP-XGB} & \textbf{CRP-XGB} \\
    \midrule
    Energy   & 86.58608 & 87.44647 & 87.39388 & 87.39582 \\
    Blog  & 23.20011 & 23.12014  & 23.08946 & 23.05300 \\
    Bank  & 0.94543 & 0.94138 & 0.94125 & 0.94164 \\
    Credict & 0.78302 & 0.78344 & 0.78345 & 0.78344 \\
    \bottomrule
    \end{tabular}%
  \label{tab:accuracy}%
\end{table}%

In general, our three secure algorithms achieve competitive accuracy against the plaintext XGB model. We conclude that the slight loss is caused by two reasons. First, we adopt fixed-point representation, with narrower precision than plaintext float-point representation. Second, secure algorithms introduce function approximation (e.g., $sigmoid$) and probable failure of $truncation$ \cite{mohassel2017secureml}.

\subsection{Efficiency Analysis on Synthetic Dataset}\label{seceff}
We analyze model efficiency by conducting the following experiments. 
The basic settings are as follows: 
\begin{itemize}
    \item The dataset is generated with 5,000 samples and 250 features.
    \item We run 10 trees with depth as 4 and feature bucket is 10.
    \item Bandwidth is 100Mbps and latency is 5ms.
\end{itemize}
In each experiment, we vary one parameter and report the average time of the main steps in Algorithm \ref{alg:trainxgb}. Unless mentioned, the above basic settings are adopted.

\begin{figure}[t]
    \centering
    \includegraphics[width=0.5\textwidth]{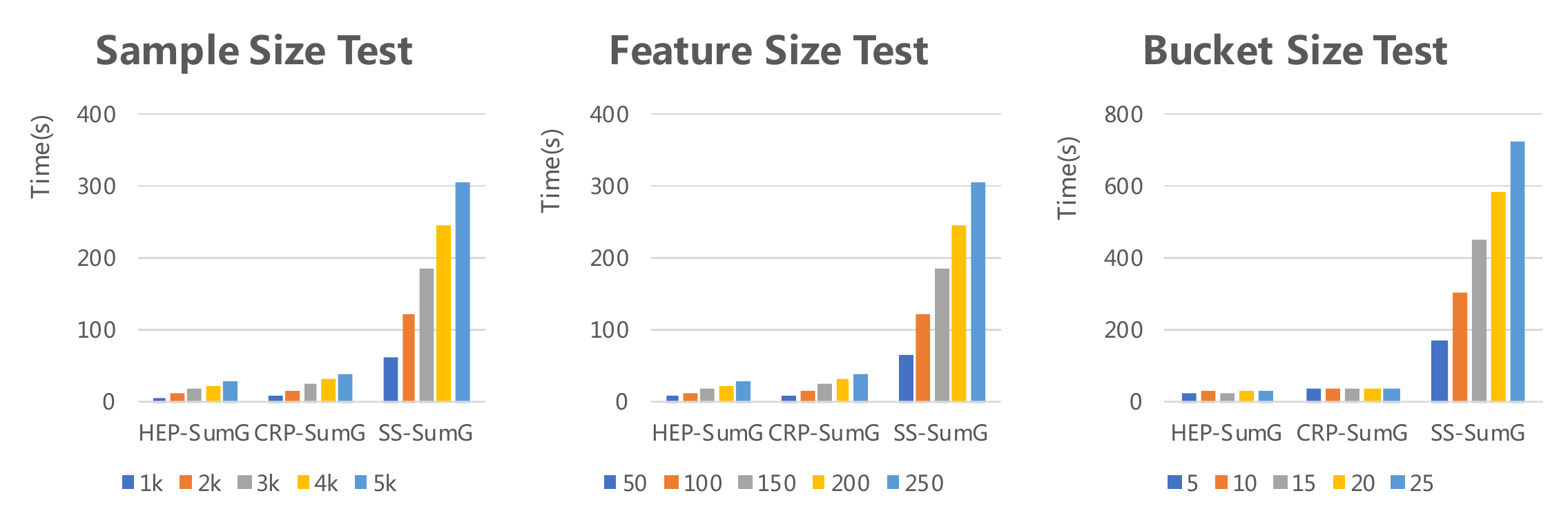}~
    \vskip -0.1in
    \caption{Effect of sample/feature/bucket size.}\label{figexp_dataset}
\end{figure}

\nosection{Effect of sample size, feature size, and bucket size}
We vary these size parameters and study their impacts on three secure XGB models. 
We report the time of $SumGradients$ for each model in Figure \ref{figexp_dataset}. 
From it, we can see that (1) the running time of each model is proportional to the scale of data, i.e., sample size and feature size; 
(2) SS-XGB spends longer time with the increase of bucket number, the same as our analysis in Section \ref{secsp};
(3) in general, the optimized two variants, HEP-XGB and CRP-XGB, obtain similar efficiency and significantly outperform SS-XGB. 
Thus, we will mainly compare HEP-XGB and CRP-XGB in the following experiments.

\begin{figure}[t]
    \centering
    \includegraphics[width=0.5\textwidth]{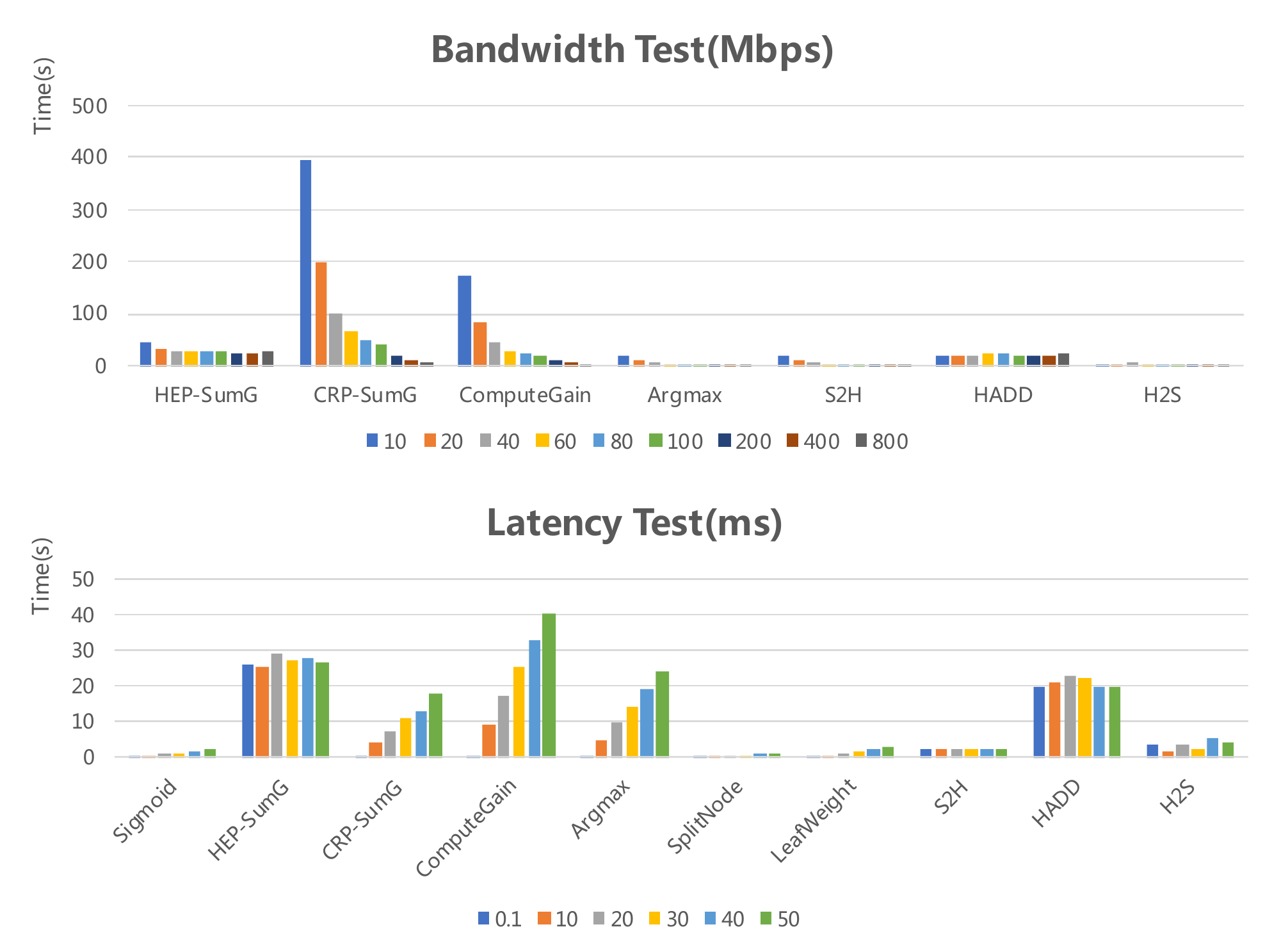}~
    \vskip -0.1in
    \caption{Effect of network ability.}\label{figexp_net}
    \vskip -0.05in
\end{figure}

\nosection{Effect of network parameters}
Because secret sharing plays a dominant role in our proposed models and it is a network sensitive scheme, we conduct detailed tests by varying the network parameters. 
To test \textit{bandwidth}, we set latency to 0 to get rid of its impact. 
Similarly, to test \textit{latency}, we use unlimited bandwidth.
We show the results in Figure \ref{figexp_net} and only report the most time-consuming steps for convenience. The $SumGradient$ implemented with HE based permutation is divided into three parts, i.e., $S2H$, $H2S$, and $HADD$. $S2H$ and $H2S$ are the transforming time cost between HE and SS, and $HADD$ denotes the time cost of accumulating the gradients under HE scheme, as described in Section \ref{secsp}.

From Figure \ref{figexp_net}, we can see that 
(1) \textit{CRP-SumG}, \textit{ComputeGain}, \textit{Argmax}, and $S2H$ are sensitive to bandwidth. 
%
Especially for \textit{CRP-SumG}, since it reveals mask permutation and mask share, which are proportional to the dataset size; 
In contrast, \textit{HEP-SumG} and its main subprocess \textit{HADD} are stable with the change of bandwidth. 
(2) with unlimited bandwidth, the time of each secret sharing process significantly decreases and CRP-based $SumGradient$ becomes more efficient than HEP-based one. 
\textit{CRP-SumG} performs the best when bandwidth is above 200 Mbps; 
(3) \textit{ComputeGain} and \textit{Argmax} are sensitive to latency, which indicates that one can optimize the round number in $Division$ and $Argmax$ to further improve efficiency.

%

\subsection{Comparison on Real-world Datasets}
We first show the accuracy comparison results in Table \ref{tab:realacc}, where the top one denotes the classification task and the bottom one denotes the regression task.  \textbf{XGB-PartyA} trains the plaintext XGB only with the features at Party A, while \textbf{XGB} uses the combined features of A and B. 
From Table \ref{tab:realacc}, we can see that 
(1) the joint models clearly improve the accuracy of XGB-PartyA, which proves the advantages of federated learning; 
(2) the accuracy of HEP-XGB and CRP-XGB is similar to XGB on the regression task but lower than XGB on the classification task.
We analyze the accuracy loss by replacing the secure $Sigmoid$ approximation with plaintext computation, and the AUC then goes up to 0.84463. 
This indicates that the approximation of $Sigmoid$ (described in Section \ref{secprot}) brings accuracy loss, especially when the dataset is large.

\begin{table}[t]
  \centering
  \caption{Accuracy comparison on real-world datasets.}
  \vspace{-0.15cm}
  \label{tab:realacc}
    \begin{tabular}{lllll}
    \toprule
    \textbf{Model} & \textbf{XGB-PartyA} & \textbf{XGB} & \textbf{HEP-XGB} & \textbf{CRP-XGB} \\
    \midrule
    \textbf{AUC} & 0.79904 & 0.84257 & 0.82945 & 0.83023 \\
    \textbf{KS} & 0.44735 & 0.52524 & 0.50270 & 0.50350 \\
    \textbf{F1 Score} & 0.20071 & 0.25190 & 0.24429 & 0.24504 \\
    \midrule
    \midrule
    \textbf{Model} & \textbf{XGB-PartyA} & \textbf{XGB} & \textbf{HEP-XGB} & \textbf{CRP-XGB} \\
    \midrule
    $R^{2}$ & 0.28634767 & 0.3324149 & 0.3312035 & 0.3320921 \\
    \textbf{RMSE} & 6459.2851 & 6246.4065 & 6253.0948 & 6248.9397 \\
    \textbf{MAE} & 2638.36573 & 2547.8378 & 2548.3600 & 2548.9299 \\
    \bottomrule
    \end{tabular}%
  \label{tab:addlabel}%
\end{table}%

We then compare the efficiency of CRP-XGB and HEP-XGB on the classification dataset under three typical environment settings:
\begin{itemize}
    \item \textbf{S1}: 32 cores, bandwidth 10 Gbps + latency 0.1 ms; 
    \item \textbf{S2}: 32 cores, bandwidth 100 Mbps + latency 5 ms; 
    \item \textbf{S3}: 8 cores, bandwidth 20 Mbps + latency 50 ms. 
\end{itemize}

We report the average training time per tree for HEP-XGB and CRP-XGB in Table \ref{tab:timetest}. 
From it, we can see that 
(1) the training time of CRP-XGB and HEP-XGB increases rapidly under bad network condition (\textbf{S3}), which indicates that both CRP-XGB and HEP-XGB rely much on the network ability; 
(2) CRP-XGB is a better choice under good network conditions. For example, CRP-XGB is about 10 times faster than HEP-XGB under \textbf{S1}; 
(3) HEP-XGB outperforms CRP-XGB under bad network conditions, e.g., HEP-XGB saves the running time by 26.5\% under \textbf{S3}.

\begin{table}[t]
  \centering
  \caption{Time test on real-world dataset}
  \label{tab:timetest}
    \begin{tabular}{llll}
    \toprule
    \textbf{Model} & \textbf{S1} & \textbf{S2} & \textbf{S3} \\
    \midrule
    \textbf{CRP-XGB} & 5.1s   & 155.8s & 857.4s \\
    \textbf{HEP-XGB} & 47.6s  & 151.0s & 630.8s \\
    \bottomrule
    \end{tabular}%
\end{table}%


\subsection{Scalability Test}
\begin{figure}[t]
    \centering
    \includegraphics[width=0.5\textwidth]{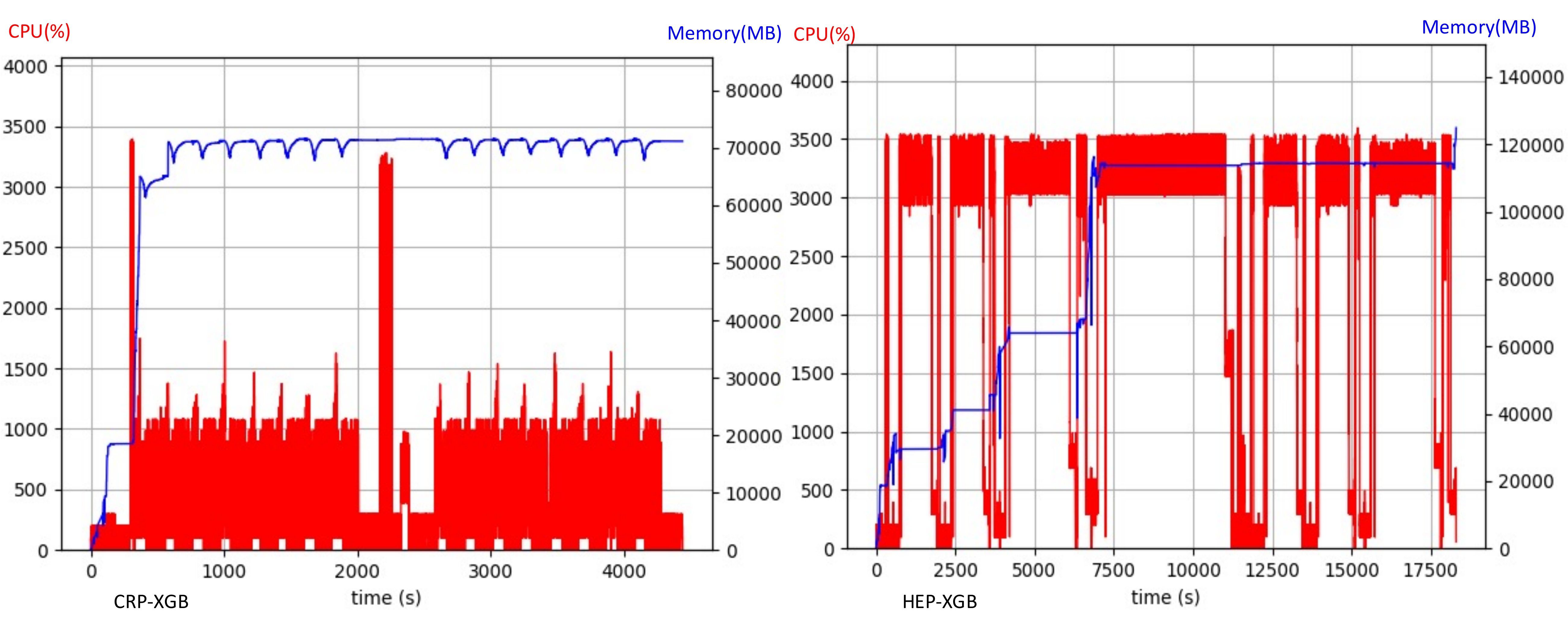}
    \vskip -0.1in
    \caption{Resource consumption. Left Figure: CRP-XGB, right figure: HEP-XGB.}\label{figexp_res}
    \vskip -0.05in
\end{figure}

To test the scalability of CRP-XGB and HEP-XGB, we compare their resource consumption based on the experimental environment reported in Section \ref{secsetup}. 
%
%
To do this, we generate a dataset with 12,000,000 samples and 200 features (considering the memory limit). We train two trees for each model and record their resource usage in Figure \ref{figexp_res}, where we use local area network. 
From it, we can see that (1) both CRP-XGB and HEP-XGB can scale to tens of millions of datasets; 
(2) in general, CRP-XGB uses much less CPU than HEP-XGB, because HEP-XGB involves time-consuming HE operations. 
The results indicate that, compared with HEP-XGB, CRP-XGB is more suitable for the situations where the network ability is good. 

\section{Conclusion and Future Work}\label{conc}
In this paper, we propose an end-to-end framework for large-scale secure XGB under vertically partitioned data setting. 
With the help of secure multi-party computation technique, we first propose SS-XGB, a secure XGB training model.
We then propose HEP-XGB and CRP-XGB, which improves the efficiency of SS-XGB based on different realizations of secure permutation. 
We also propose a secure XGB prediction algorithm. 
The experimental results show that our proposed models achieve competitive accuracy with the plaintext XGB and can scale to large datasets.
In the future, we plan to further optimize the communication of the secret sharing protocols, which is the main bottleneck in our industrial test. 

\bibliographystyle{ACM-Reference-Format}
\bibliography{hessxgb}
\end{document}